\documentclass[lettersize,journal]{IEEEtran}
\usepackage{amsmath,amsfonts}
\usepackage{algorithmic}
\usepackage{array}
\usepackage[caption=false,font=normalsize,labelfont=sf,textfont=sf]{subfig}
\usepackage{textcomp}
\usepackage{stfloats}
\usepackage{url}
\usepackage{verbatim}
\usepackage{graphicx}
\usepackage{float}
\usepackage{booktabs}
\def\BibTeX{{\rm B\kern-.05em{\sc i\kern-.025em b}\kern-.08em
    T\kern-.1667em\lower.7ex\hbox{E}\kern-.125emX}}
\usepackage{balance}
\begin{document}
\title{BEVStereo++: Accurate Depth Estimation in Multi-view 3D Object Detection via Dynamic Temporal Stereo}
\author{
    Yinhao Li, Jinrong Yang, Jianjian Sun, Han Bao, Zheng Ge, Li Xiao*
   \thanks{Li Xiao is  with  School of Artificial Intelligence, Beijing University of Posts and
Telecommunications, Beijing, China. Li Xiao is the corresponding author. Email: andrew.lxiao@gmail.com, andrewxiao@bupt.edu.cn. }
\thanks{Yinhao Li and Han Bao are with Institute of Computing Technology, Chinese Academy of Sciences, Beijing, China.}
\thanks{Jinrong Yang is with Huazhong University of Science and Technology, Wuhan, China.}
\thanks{Jianjian Sun and Zheng Ge are with MEGVII Technology, Beijing, China.}
}


\maketitle

\begin{abstract}
Bounded by the inherent ambiguity of depth perception, contemporary multi-view 3D object detection methods fall into the performance bottleneck. Intuitively, leveraging temporal multi-view stereo (MVS) technology is the natural knowledge for tackling this ambiguity. However, traditional attempts of MVS has two limitations when applying to 3D object detection scenes: 1) The affinity measurement among all views suffers expensive computational cost; 2) It is difficult to deal with outdoor scenarios where objects are often mobile. To this end, we propose BEVStereo++: by introducing a dynamic temporal stereo strategy, BEVStereo++ is able to cut down the harm that is brought by introducing temporal stereo when dealing with those two scenarios. Going one step further, we apply Motion Compensation Module and long sequence Frame Fusion to BEVStereo++, which shows further performance boosting and error reduction. Without bells and whistles, BEVStereo++ achieves state-of-the-art(SOTA) on both Waymo and nuScenes dataset.
\end{abstract}

\begin{IEEEkeywords}
multi-view, 3D object detection, motion compensation, dynamic temporal stereo, memory efficiency
\end{IEEEkeywords}

\section{Introduction}
\IEEEPARstart{D}{ue} to the stability and inexpensive cost of vision sensors, camera-based 3D object detection has received extensive concern. Specially, the camera-based schemes~\cite{wang2022detr3d, huang2021bevdet, liu2022petr, li2022bevformer, huang2022bevdet4d, liu2022petrv2, li2022bevdepth, liu2021yolostereo3d, li2019stereo} show significantly promising, and have made lots of breakthroughs. However, there is still a substantial performance gap compared with LiDAR-based approaches~\cite{lang2019pointpillars, yan2018second, yin2021center, dbqssd}, since it exposes a notoriously ill-posed issue for perceiving depth.

Contemporary multi-view detectors~\cite{huang2021bevdet, huang2022bevdet4d, li2022bevdepth} predict a discrete depth distribution for each point of the field of view (FOV), which enables to project features from image representation to Bird’s Eye View(BEV) map. The unified BEV map is the key to learn harmonious results since the overlap regions of adjacent views represent more complete to directly forecast results. Such sweetness is hard to be enjoyed by the monocular-based detector~\cite{wang2021fcos3d}, as a post-processing strategy is needed to remove repetitive and low-quality 3D boxes in overlap areas. 

The above paradigm is based on an important preassumption, i.e., the perceived depth distribution in FOV needs to be accurate enough. However, most perceived depth is obtained by only feeding into single-frame images, which is an ill-posed solution~\cite{huang2021bevdet, huang2022bevdet4d, li2022bevdepth}. Several studies~\cite{yao2018mvsnet, xue2019mvscrf, bae2022multi} point out that predicting depth needs \textbf{multi-view stereo(MVS)} condition, which requires images from different views to construct temporal stereo. Fortunately, the automatic driving scenario is often processed in a continuous time sequence, enabling us to leverage temporal views for constructing multi-view stereo.

To carry out the traditional temporal stereo technology such as \cite{yao2018mvsnet} is non-trivial in automatic driving scenarios, which manifests in two aspects:

\begin{enumerate}
\item \textbf{Large memory cost:} taking BEVDepth~\cite{ li2022bevdepth} as an example, 
 although BEVDepth achieves the previous state-of-the-art(SOTA) results by proposing a reliable depth prediction module, it can only performs depth prediction on single frame images. On the other hand, When we replace the depth module in BEVDepth with ~\cite{yao2018mvsnet}'s temporal stereo approach, there comes another issue that the memory cost increases by 3.5 times with only 1.6 percent promotion on NDS, posing a significant strain in real-world applications;

\item \textbf{Failing at key scenarios:} temporal stereo approaches are unable to handle several key tasks~\cite{wang2022monocular} such as reasoning the depth of moving objects and static ego vehicle cases, since the parallax angle tends to 0 if ego vehicle is static and the stereo is unable to match if the object is moving. However, the statistics on the nuScenes dataset~\cite{caesar2020nuscenes} show that over 10\% of the frames' ego vehicles are static, and approximately 25\% of the objects are moving. 
\end{enumerate}

The above two shortcomings limit its application to autonomous driving scenarios. Because the working function of MVS is to find the best matching points of two frames by constructing temporal stereo, introducing scenarios such as moving objects and static ego vehicles will jeopardize the MVS method's foundation. Furthermore, as MVS methods construct temporal stereo along the entire depth axis, the memory cost of applying temporal stereo to the current multi-view 3D object detection task has become unacceptably high. To this end, we propose a dynamic temporal stereo technique for correcting the two flaws of MVS-based methods while retaining the high quality of depth prediction provided by temporal stereo construction. By introducing the parameters $\mu$(depth center) and $\sigma$(depth range) to dynamically construct temporal stereo rather than constructing all candidates along the depth axis at once, the model is able to focus on the candidates with the highest probability while maintaining memory efficiency. Going one more step, we introduce a parameter evolution method for $\mu$ and $\sigma$, which is carried out by applying the EM algorithm to update the modeling parameters $\mu$ and $\sigma$. When confronted with scenarios that MVS methods can handle, the EM algorithm allows $\mu$ and $\sigma$ to approach the depth GT while remaining unchanged for other scenarios. Furthermore, we incorporate Motion Compensation Module into our method, which is based on the dynamic temporal stereo, to obtain related pixels of two frames even when objects are moving. Finally, we also introduce an advanced variant of Circle NMS~\cite{yin2021center}, which considers objects' size for better removing duplicate 3D boxes.

 A preliminary work named as \textbf{BEVStereo} has been accepted by AAAI 2023~\cite{li2022bevstereo}, which proposed a dynamic temporal stereo technique to dynamically select the scale of matching candidates. In this paper, we extend this work to \textbf{BEVStereo++}, which enjoys the beauty of temporal stereo while avoiding incidental drawbacks brought by temporal stereo. By conducting comprehensive experiments on nuScences~\cite{caesar2020nuscenes} and Waymo~\cite{sun2020scalability} benchmark, It achieves success in the key scenarios and makes great progress in the 3D object detection task. The main contributions of BEVStereo++ can be summarized as the following three aspects:

\begin{itemize}
\item We propose a dynamic temporal stereo strategy. When compared to existing MVS-based approaches, our method provides higher performance with reduced memory consumption by exploiting the depth center ($\mu$) and depth range ($\sigma$) to generate temporal stereo dynamically.

\item We design a Motion Compensation Module to address the moving object issue. With the help of 
EM algorithm to update $\mu$ and $\sigma$, our method avoids failure at key scenarios(moving objects and static ego vehicle). We also propose Frame Fusion, Size-aware Circle NMS, Efficient Voxel Pooling modules to further improve the efficiency and accuracy of our method.

\item With the assistance of the new design, BEVStereo++ improves mAP and NDS by 3.2\% and 3\% on nuScenes dataset than the previous version of our work-BEVStereo, achieving the new SOTA performance on the camera-only track while maintaining high efficiency. 
\end{itemize}

\section{Related Work}

\subsection{Single-view 3D Object Detection}
Many approaches have made their effort on predicting objects directly from single images. For the purpose of 3D object detection, Cai et al.~\cite{cai2020monocular} calculates the depth of the objects by integrating the height of the objects in the image with the height of the objects in the real world. Based on FCOS~\cite{tian2019fcos}, FCOS3D~\cite{wang2021fcos3d} extends it to 3D object detection by changing the  classification branch and regression branch which predicts 2D and 3D attributes at the same time. M3D-RPN~\cite{brazil2019m3d} treats mono-view 3D object detection task as a stand-alone 3D region proposal network, narrowing the gap between LiDAR-based approaches and camera-based methods. D$^4$LCN~\cite{ding2020learning} replaces 2D depth map with pseudo LiDAR representation to better present 3D structure. DFM~\cite{wang2022monocular} integrates temporal stereo to mono-view 3D object recognition, improving the quality of depth estimation while minimizing the negative effects of difficult situations that temporal stereo is unable to handle.
\subsection{Multi-view 3D Object Detection}
Current multi-view 3D object detectors can be divided into two schemas: LSS-based ~\cite{philion2020lift} schema and transformer-based schema. 

BEVDet ~\cite{huang2021bevdet} is the first study that combines LSS and LiDAR detection head which uses LSS to extract BEV feature and uses LiDAR detection head to propose 3D bounding boxes. By introducing previous frames, BEVDet4D~\cite{huang2022bevdet4d} acquires the ability of velocity prediction. To reduce memory usage, M$^2$BEV~\cite{xie2022m}  decreases the learnable parameters and achieves high efficiency on both inference speed and memory usage. BEVDepth~\cite{li2022bevdepth} uses LiDAR to generate depth GT for supervision and encodes camera intrinsic and extrinsic parameters to enhance the model's ability of depth perception.

DETR3D~\cite{wang2022detr3d} extends DETR ~\cite{carion2020end} into 3D space, using transformer to generate 3D bounding boxes. Based on DETR, PETR~\cite{liu2022petr} and PETRV2~\cite{liu2022petrv2} adds position embedding onto it. BEVFormer~\cite{li2022bevformer} uses deformable transformer to extract features from images and uses cross attention to link the feature between frames for velocity prediction. STS~\cite{wang2022sts} is the first multi-view 3D object detector that used temporal stereo technique and achieved SOTA on the nuScenes dataset.

\subsection{Depth Estimation}
Based on the number of images used for depth estimation, depth estimation methods can be divided into single-view depth estimation and multi-view depth estimation.

Although predicting depth from a single image is obviously ill-posed, it is still possible to estimate some of the depth of the objects by using the context as a signal. Therefore, many approaches~\cite{bhat2021adabins, eigen2015predicting, eigen2014depth, fu2018deep} use CNN method to predict depth.

For the task of multi-view depth estimation, Constructing temporal stereo is an effective way to predict depth~\cite{mvs_survery, aa_rmvsnet}. MVSNet~\cite{yao2018mvsnet} is the first research that uses temporal stereo for depth estimation. RMVSNet~\cite{yao2019recurrent} reduces memory cost by introducing GRU module. MVSCRF~\cite{xue2019mvscrf} adds CRF module onto MVSNet. PointMVSNet~\cite{chen2019point} uses point algorithm to optimize the regression of depth estimation. Cascade MVSNet~\cite{gu2020cascade} uses cascade structure, making it able to use large depth range and a small amount of depth intervals. Fast-MVSNet~\cite{yu2020fast} uses sparse temporal stereo and Gauss-Newton layer to speed up MVSNet. Wang et al.~\cite{wang2021patchmatchnet} use adaptive patchmatch and multi-scale fusion to achieve good performance while mataining high efficiency. Bae et al.~\cite{bae2022multi} introduce MaGNet to better fuse single-view depth estimation and multi-view depth estimation.
 
\section{Method}
BEVStereo++ is a stereo-based multi-view 3D object detector.
By applying the temporal stereo technique with Motion Compensation Module, our method is able to handle complex outdoor scenarios while maintaining memory efficiency. In addition, BEVStereo++ is applied with long sequence Frame Fusion design to construct temporal stereo which shows superiority. Finally, a size-aware circle NMS approach is proposed to improve the proposal suppression process.

\subsection{Preliminary Knowledge}
\paragraph{Multi-view 3D object detection}
LSS-based~\cite{philion2020lift} multi-view 3D object detectors currently include four components: an image encoder to extract the image features, a depth module to generate depth and context, and then outer product them to obtain point features, a view transformer to convert the feature from camera view to the BEV view, and a 3D detection head to propose the final 3D bounding boxes.
\paragraph{Temporal stereo methods to predict depth}
MVS-based~\cite{yao2018mvsnet} methods predict depth by constructing temporal stereo. For every pixel on the reference feature, they initially put forth a number of candidates along the depth axis. Next, they convert these candidates from reference to source using a homography warping operation in order to retrieve the relevant source feature and create the temporal stereo. After temporal stereo is constructed. For the purpose of predicting the confidence of each depth candidate, 3D convolution is performed to regularize the temporal stereo.

\subsection{Dynamic Temporal Stereo}
In the MVS-based methods~\cite{wang2022sts, wang2022monocular}, because all candidates along the depth axis are needed to build temporal stereo, it has become the most expensive computational cost. To address this, we provide dynamic temporal stereo, which constructs temporal stereo by applying a dynamic range along the depth axis. To achieve that, we introduce $\mu$ and $\sigma$, which indicate the dynamic temporal stereo's center and length, respectively. With our design, we can improve depth estimate quality while incurring minimal computing cost.

\begin{figure*}[t]
\includegraphics[width=0.95\textwidth]{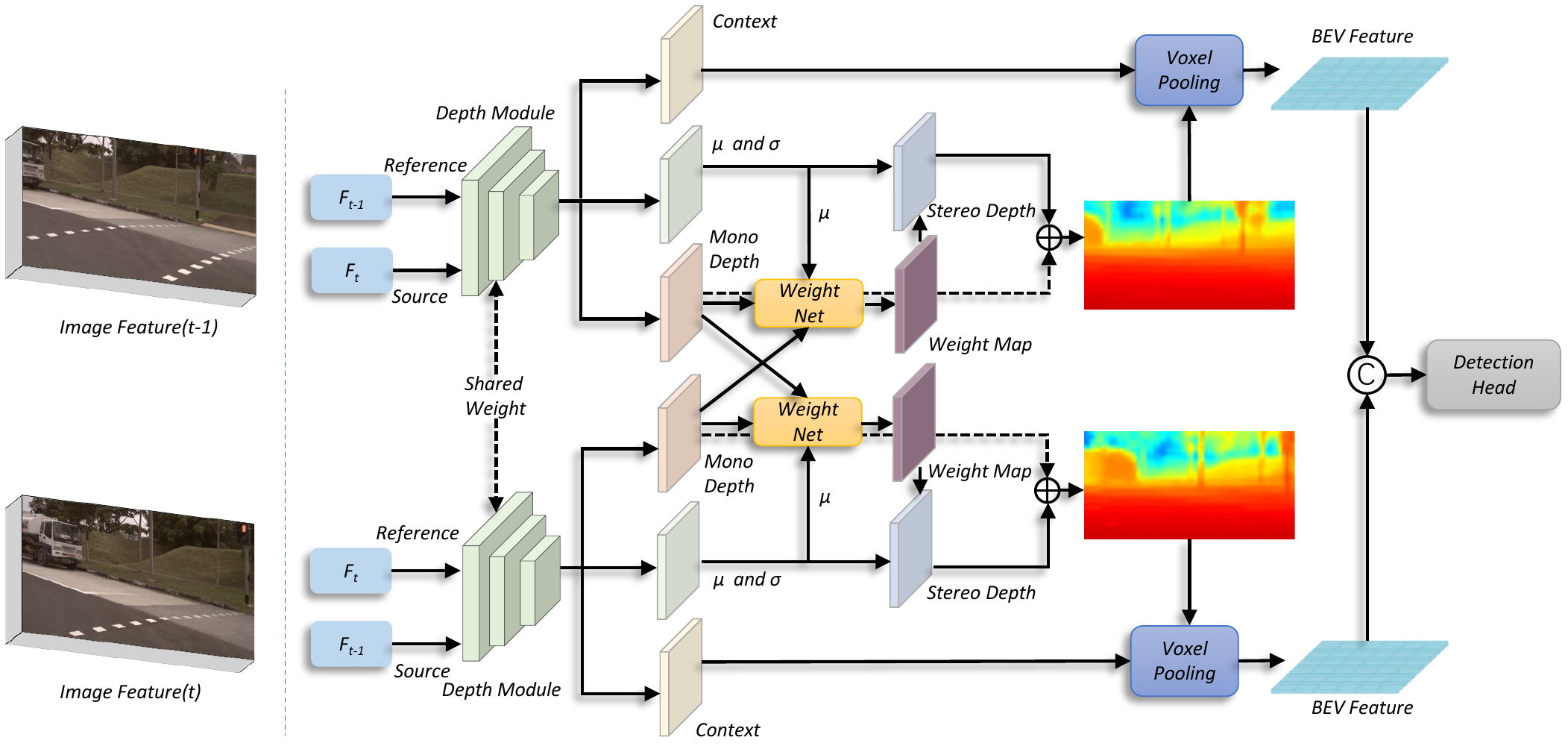}
\centering
\caption{Framework of BEVStereo++. The Depth Module uses the image feature of the reference frame and source frame as input to generate $\mu$, $\sigma$, context, and mono depth. Stereo depth is produced using $\mu$ and $\sigma$. Weight Net uses $\mu$ and the mono depth of two frames to create a weight map that is applied to the stereo depth. Mono depth and weighted stereo depth are accumulated together to create the final depth. BEV Feature is produced when context is combined with it and is used by the detecting head.}
\label{fig:bevdepth}
\end{figure*}
Inspired by DFM~\cite{wang2022monocular}, BEVStereo++ predicts depth from a single feature (mono depth) as well as depth from temporal stereo (stereo depth). Our Depth Net predicts mono depth, $\mu$, and $\sigma$ all at once. After the EM algorithm iterates $\mu$ and $\sigma$, $\mu$ and $\sigma$ are used to generate stereo depth. Weight Net is used to combine mono depth and stereo depth to generate the final depth prediction. We use motion compensation on the image features used to construct temporal stereo to improve the quality of depth prediction in outdoor scenarios. Our framework overview is illustrated in Fig.~\ref{fig:bevdepth}. 

\paragraph{Depth Module}
\begin{figure*}[t]
\includegraphics[width=0.95\textwidth]{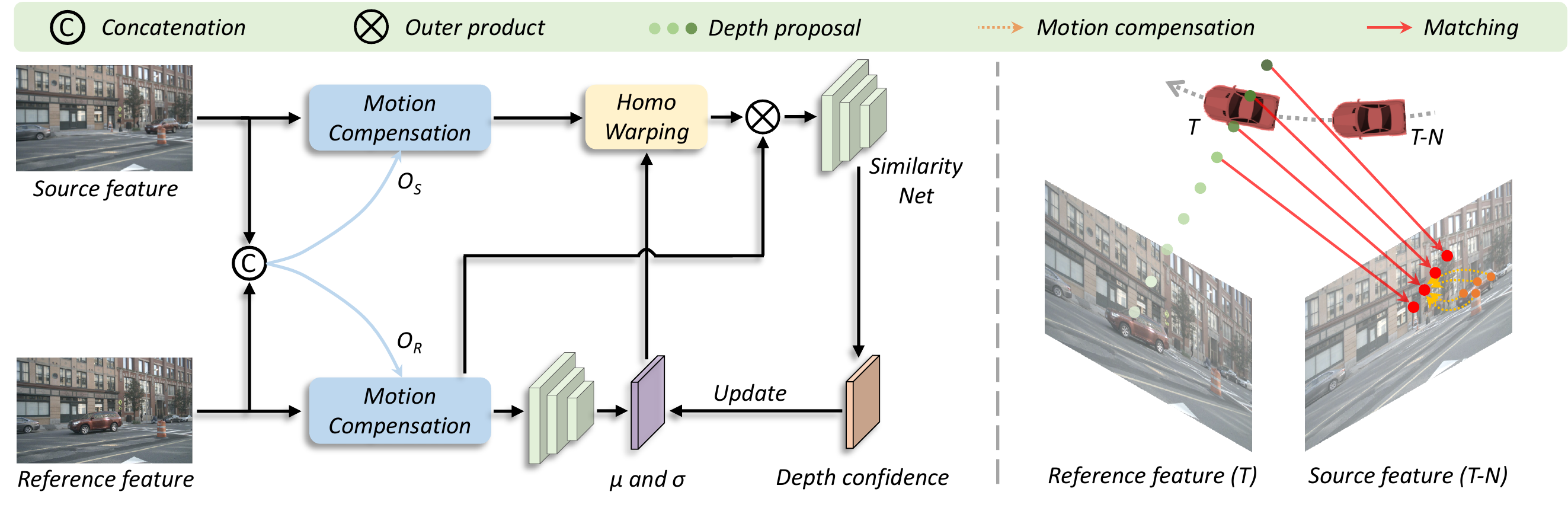}
\centering
\caption{Framework of dynamic temporal stereo. The source feature and reference feature are first concatenated to obtain motion compensation for both features, and then generate the initial $\mu$ and $\sigma$ using feature of the reference frame as input. For each round of iteration, $\mu$ and $\sigma$ are used for homography warping to fetch the source feature, and the Similarity Net takes the inner product results of warpped source feature and reference feature as input to generate depth confidence which is then used to update $\mu$ and $\sigma$.}
\label{fig:depth_module}
\end{figure*}
Our Depth Module simultaneously predicts  mono depth, $\mu$, $\sigma$ and context. After iterating $\mu$ and $\sigma$ by our EM method, they are used to generate the stereo depth. The process of iterating $\mu$ and $\sigma$ is illustrated in Fig.~\ref{fig:depth_module}.

To guarantee that the initial value of $\mu$ and $\sigma$ are not far from the depth GT, both of them are predicted by Depth Module which takes image feature of single frame as input. By doing so, we can ensure that the starting values of $\mu$ and $\sigma$ are of the same quality as the monocular depth estimation model.Compared to other stereo-based methods of splitting bins along the depth dimension~\cite{yao2018mvsnet, wang2022mv}, our method can dynamically choose the search area while also lowering the number of candidates. 

After estimating $\mu$ and $\sigma$ of the reference frame, the depth candidates for each pixel are able to be selected based on the value of $\mu$ and $\sigma$. These candidates are used for homography warping operation to fetch the feature from source frame, as illustrated in Equ.~\ref{equ: homowarping}, where $P$ denotes the coordinate of the point, $D$ denotes the depth of the candidate, $src$ denotes source frame, $ref$ denotes reference frame, $M_{ref2src}$ denotes the transformation matrix from the reference frame to source frame and $K$ denotes the intrinsic matrix. The reference feature and the warpped source feature are used to construct temporal stereo. Similarity Net is followed to predict the confidence score of all candidates.

\begin{equation}
\label{equ: homowarping}
    P_{src}[u\cdot z, v \cdot z, z] = K \times M_{ref2src} \times K^{-1} \times (D \cdot P_{ref}[u, v, 1])
\end{equation}
\begin{equation}
\label{equ: weight-sum}
    \mu = \sum_{i=1}^n D_i \cdot P_i,
\end{equation}

\begin{equation}
\label{equ: update-sigma}
    \sigma_{new} =  \frac{\sigma_{old}}{2 \cdot P_{\mu}},
\end{equation}

Inspired by the EM algorithm, we seek to bring the expectation of $\mu$ closer to the depth GT during the iteration phase. Because each pixel in the reference frame has several candidates along the depth channel, and the scores of all the candidates are computed using Similarity Net. It is only logical that we would apply this knowledge to further our goals. As a result, we apply the weight sum approach to update $\mu$. The update rule is illustrated in Equ.~\ref{equ: weight-sum}, where $D_i$ denotes the depth of the $i$th candidate and $P_i$ denotes the probability of the $i$th candidate. As $\mu$ is being updated in the process of iteration, it is also critical to find the suitable $\sigma$ to set the searching range. In accordance with existing information, the searching range should be reduced when the confidence of $\mu$ is high and expanded when it is low, we update $\sigma$ following Equ.~\ref{equ: update-sigma} where $P_{\mu}$ denotes the confidence of $\mu$. Without introducing any learnable parameters, both $\mu$ and $\sigma$ will be adapted to the change of camera positions and  the search range is optimized during iteration.

\begin{equation}
\label{equ: depth-map}
    P = exp(-\frac{1}{2} \cdot (\frac{D - \mu}{\sqrt{\sigma}})^2).
\end{equation}

To prevent the scenario where the projected $\mu$ is far from the depth GT, making it difficult to optimize $\mu$ during iteration. we divide the depth into different ranges and use our iteration technique in each split range. After the iteration process is finished, the depth map is generated following  Equ.~\ref{equ: depth-map} where P denotes the computed depth confidence and D denotes the depth of the split bins along the depth axis for each pixel.

\paragraph{Weight Net}
Although depth module is able to generate mono depth and stereo depth, how to combine them to get the final depth prediction remains unsolved. Considering not all pixels of stereo depth is reliable since not all pixels have related point in the source frame, we apply Weight Net to generate pixel-wise weight map that is used to fuse mono depth and stereo depth. To achieve that, we use the final $\mu$ as depth to transform the mono depth from two frames onto the same plane, and the weight map is generated using the mono depth from two frames.

\paragraph{Motion Compensation Module}
All of the designs mentioned above is able to prevent our model from failing when dealing with complex outdoor scenarios. However, when confronted with moving objects, our model is still unable to use temporal stereo to improve depth prediction quality because $\mu$ is generated by a single frame and tends to remain unchanged when objects are moving. To this end, As shown in Fig.~\ref{fig:depth_module}, we design a motion compensation module to improve the accuracy of moving objects depth estimation. We adopt a deformable convolution structure, which can forecast the offsets of the convolution kernel and acquire features based on the expected offsets, and compensate for the change caused by motion. Specifically, we use the concatenated source feature and reference feature as input to predict the offsets for both features, and then the predicted offsets are applied on each feature to complete the motion compensation process.

As shown in the right part of Fig.~\ref{fig:depth_module}, we take one point from the reference feature as an example. The proposed points initially are unable to reach the related point in the source feature. The Motion Compensation Module is able to adjust the proposed points to reach the correct point in the source domain.

\begin{figure}[t]
\centering
\includegraphics[width=0.48\textwidth]{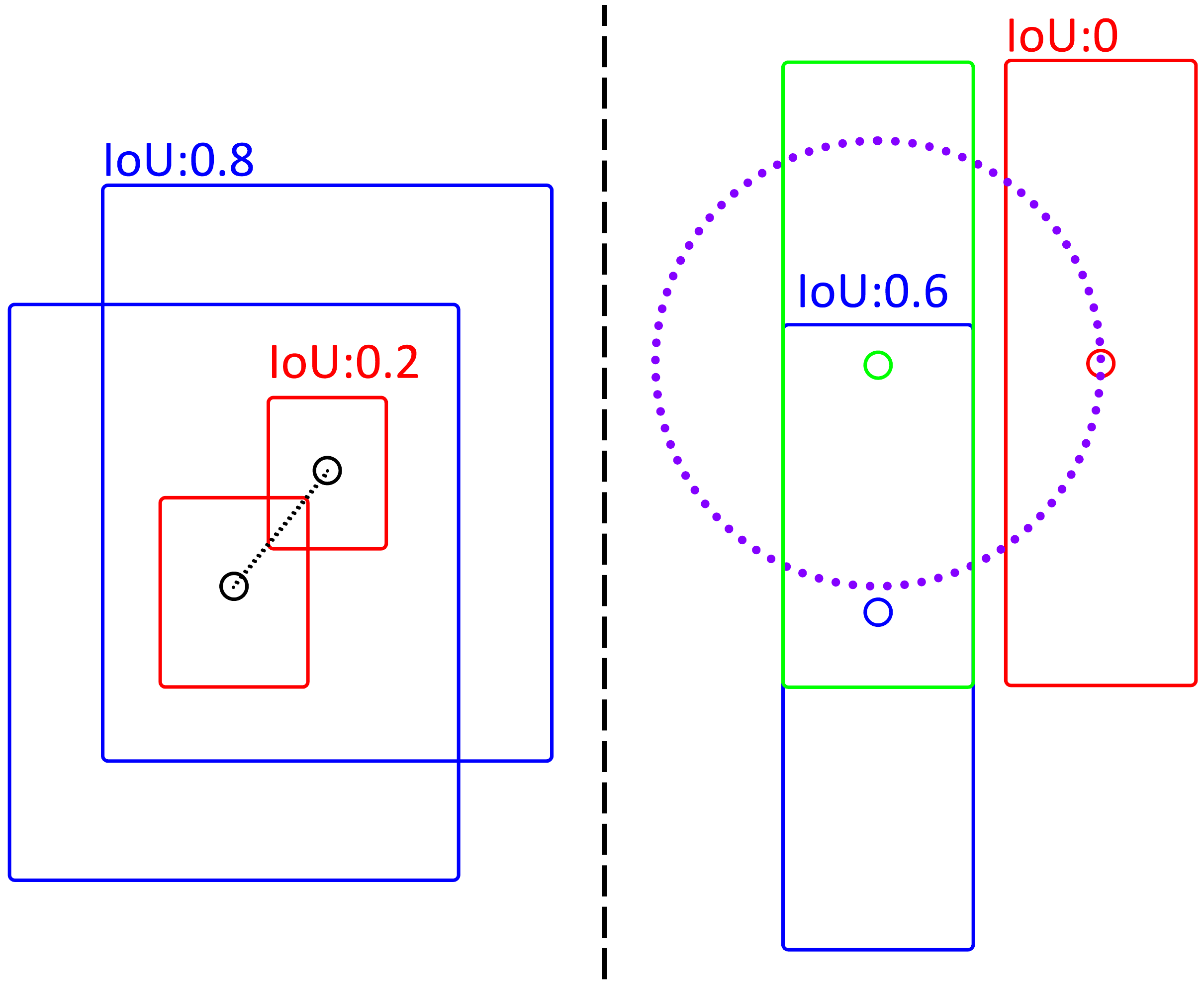}
\centering
\caption{Illustration of our frame fusion schema. For dynamic temporal stereo, two adjacent frames are used, but only one BEV frame is created. Following the point alignment module, BEV features from previous frames are concatenated with BEV features from the current frame.}
\label{fig:long_sequence}
\end{figure}

\begin{figure}[t]
\centering
\includegraphics[width=0.48\textwidth]{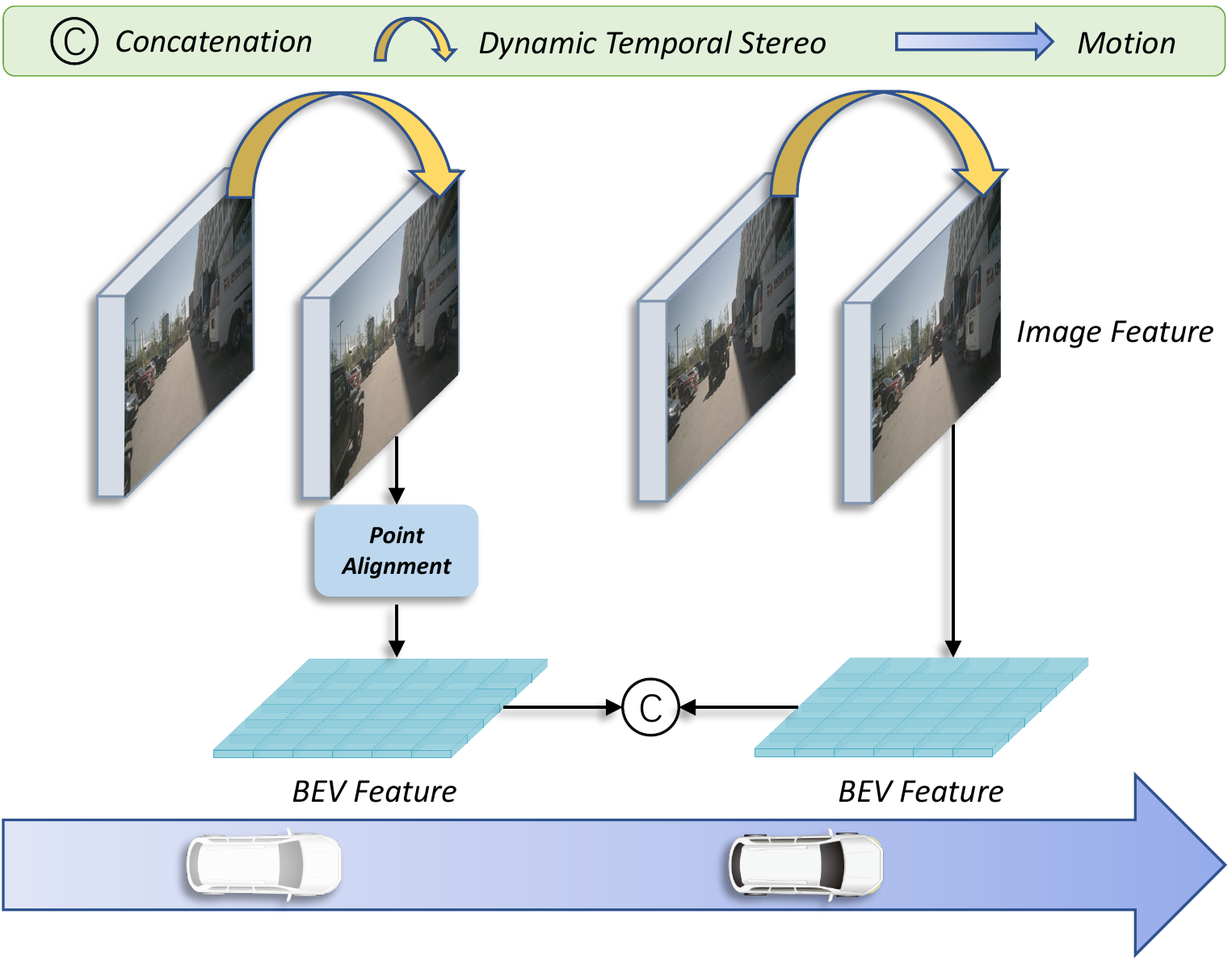}
\centering
\caption{Drawbacks of the circle NMS: On the left side of the figure, despite having distinct IoUs, the blue boxes(IOU=0.8) and red boxes(IOU=0.2) share the same center distance and surrend to the same suppression criterion. In the right part of the figure, although the objects in the green box and red box are non-overlap(IOU=0),they may suppress each other as their center distance is within the suppression threshold.}
\label{fig:circle-nms}
\end{figure}

\subsection{Frame Fusion}

As SOLOFusion~\cite{park2022time} points out, a long sequence improves the model's perception ability. We also incorporate long sequence design into our model. As MVS~\cite{yao2018mvsnet} demonstrates, short term and high resolution are good for constructing temporal stereo so that related points can be better matched, while long term is useful for supplementing blind areas. In this work, instead of using different resolutions to construct temporal stereo on both the short and long term, we design a sliding window schema for long term and short term fusion.

As illustrated in Fig.\ref{fig:long_sequence}, we use two types of frame fusion schema - image feature fusion and BEV feature fusion. We sample four frames from the timeline and divide them into two groups, with adjacent two frames assigned to each group. Image feature fusion is utilized to generate BEV features within each group. All BEV features are fused together after the Point Alignment Module to form the final BEV feature.

To fuse image features within each group, we apply the dynamic temporal stereo method. Instead of building dynamic temporal stereo for all frames inside each group, we only construct it for the most recent frame within each group, treating it as a reference frame and the other frames as source frames. By doing that, we are able to avoid the time delay caused by constructing dynamic temporal stereo.

\begin{equation}
\label{eq:multi}
\mathcal{P}^{\mathrm{cur}}=\mathcal{T}^{\mathrm{global2cur}} \cdot  \mathcal{T}^{\mathrm{prev2global}} \cdot \mathcal{P}^{\mathrm{prev}}.
\end{equation}

Since our method generates pseudo LiDAR points to lift image features into 3D space, it is natural to employ the pseudo LiDAR points to facilitate the process of BEV feature fusion rather than the grid sample strategy. To achieve that, we employ the Point Alignment Module, which transforms the pseudo LiDAR points from previous frames into the current frame by Equ.~\ref{eq:multi}, where $\mathcal{T}$ denotes translation-rotation matrix. Following the point alignment procedure, image features from all frames can be mapped into BEV features using the Voxel Pooling v2. Our designed frame fusion is more accurate as we fuse the information in 3D space (LiDAR points) rather than 2D space (BEV feature), and also has less computational cost since it avoids the BEV feature warp operation.

\subsection{Size-aware Circle NMS}
The bounding boxes in 3D object detection tasks are rotated rather than aligned boxes. The standard NMS approach has become a burden for the 3D object detection pipeline due to the high cost of computing IoU between rotated boxes. To address this issue, ~\cite{yin2021center} propose circle NMS, which employs the distance between the centers of two bounding boxes as a suppression criterion. Circle NMS achieves excellent efficiency and high performance by bypassing the difficult process of computing rotated IoU of bounding boxes. However, circle NMS also ignores the size of boxes, which will result in two drawbacks as illustrated in Fig.~\ref{fig:circle-nms}: 1) the suppression criterion only depends on box center distances, large objects are likely to generate multiple predicted bounding boxes as highly overlapped proposals may not be suppressed; 2) small objects are likely to be ignored as non-overlapping proposals may still be removed.

To this end, we propose the size-aware circle NMS to avoid computing rotated IoU while considering the boxes' size. Like the circle NMS, our method keeps the box center points distance as the suppression criterion. Differently, we divide the suppression criteria into the distances between the box center boxes at the x-axis and y-axis separately, and the criteria value depends on the box sizes. We use $x_{thre}$ and $y_{thre}$ to represent the suppression thresholds of the x-axis and y-axis, computed following Equ.~\ref{equ: xthre} and Equ.~\ref{equ: ythre}. Here $\theta_1$ and $\theta_2$ denote the two box orientations, $w$ denotes the hyperparameter of scale factor, $d_x$ denotes the length of the box, and $d_y$ denotes the width of the box. Suppression occurs when the box center distance in the x-axis is smaller than $x_{thre}$ and the box center distance in the y-axis is smaller than $y_{thre}$. By applying size-aware circle NMS, the blue boxes on the left side of Fig.~\ref{fig:circle-nms} will trigger suppression as it obtains greater $x_{thre}$ and $y_{thre}$. Suppression will not occur on the right side of  Fig.~\ref{fig:circle-nms} because the distances in the x and y axes are more likely to be smaller than $x_{thre}$ and $y_{thre}$ in the meantime.
\begin{equation}
\label{equ: xthre}
    x_{thre} = w \cdot (cos\theta_1 \cdot d_{x1} + sin\theta_1 \cdot d_{y1} + cos\theta_2 \cdot d_{x2} + sin\theta_2 \cdot d_{y2}).
\end{equation}

\begin{equation}
\label{equ: ythre}
    y_{thre} = w \cdot (cos\theta_1 \times d_{y1} + sin\theta_1 \cdot d_{x1} + cos\theta_2 \cdot d_{y2} + sin\theta_2 \cdot d_{x2}).
\end{equation}

\begin{figure}[htbp]
\centering
\includegraphics[width=0.48\textwidth]{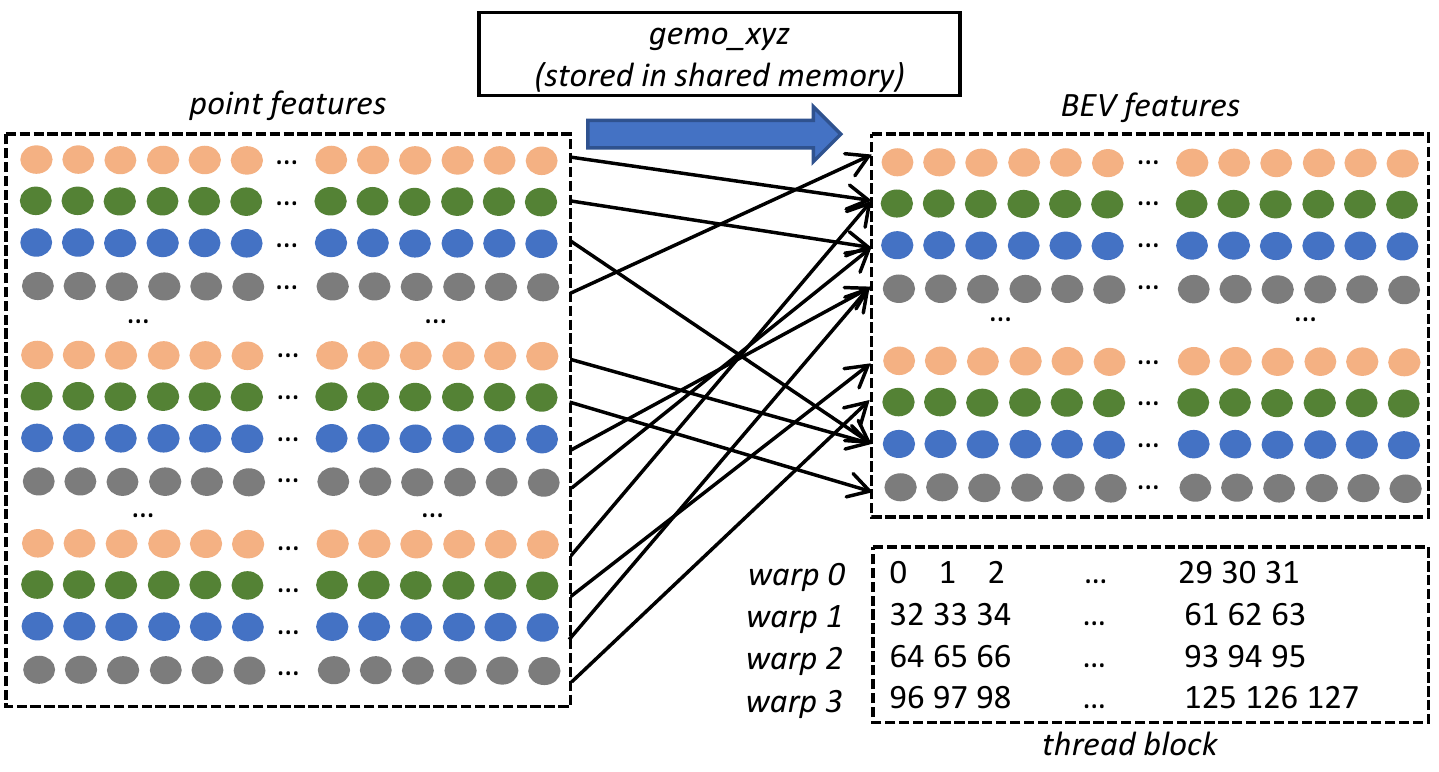}
\centering
\caption{Thread mapping of point features to BEV features. Based on the point coordinates, the point features are automatically accumulated into the corresponding BEV features. Each thread block loads the point coordinates it is responsible for into the shared memory.}
\label{fig:mapping}
\end{figure}

\subsection{Efficient Voxel Pooing v2}

In the previous version of Efficient Voxel Pooling~\cite{li2022bevdepth}, threads within the same warp access memory discontinuously, leading to more memory transactions, which results in poor performance. We enhance Efficient Voxel Pooling by improving the way threads are mapped, as illustrated in Fig.~\ref{fig:mapping}. For each block, we employ 32 and 4 threads on the x and y axes. First, 128 point coordinates are loaded into shared memory by all the threads in one block. Then, one point feature at a time is processed by each warp. According to the point coordinates, the point feature is automically accumulated to the matching BEV feature. The 128 point features are processed round robin by four warps in a block till they are finished. In this manner, performance-limiting memory transactions from the L2 cache and global memory are diminished.

We discover that operations can be fused in the existing voxel pooling architecture inspired by BEVPoolv2~\cite{huang2022bevpoolv2}. In Efficient Voxel Pooling v1, the depth feature and the context feature are outer producted to create features of the pseudo-LiDAR points, which are then used for pooling. The implementation of voxel pooling is made simpler by creating features from pseudos LiDAR points. However, the outer product consumes high computational costs. In this work, instead of creating fake LiDAR features, we collect the corresponding features using depth and context features directly to avoid loss of information caused by outer-product. Specifically, we set up each thread to gather the source features from the depth and context features, multiply them, and then add the target feature in the appropriate place.

\section{Experiment}
In this section, we first describe the experimental settings that we employ before going into the specifics of our implementation strategy. Experiments involving several ablations are carried out to confirm the efficiency and validity of BEVStereo++.

\subsection{Experimental Settings}
\paragraph{Dataset and evaluation metrics}
We decide to run our experiments on the nuScenes~\cite{caesar2020nuscenes} and Waymo~\cite{sun2020scalability} open dataset. For training, we use LiDAR and image data, but we only use image data for inference. In the case of image data, the key frame image and the furthest sweep connected to it are used, whereas in the case of LiDAR data, only the key frame data is used. We assess the results of our method using detection and depth metrics. Memory usage is also used as a measure of the effectiveness of our methods. To be more specific, for nuScenes dataset, we report the mean Average Precision (mAP), nuScenes Detection Score (NDS), mean Average Translation Error (mATE), mean Average Scale Error (mASE), mean Average Orientation Error (mAOE), mean Average Velocity Error (mAVE), and mean Average Attribute Error (mAAE); for Waymo open dataset, we report mean Average Precision(mAP), mean Average Precision Weighted by Heading(mAPH) and mean Longitudinal Affinity Weighted LET-3D-AP(mAPL). We follow the established evaluation procedures for the depth estimation task~\cite{deptheval}, reporting scale invariant logarithmic error (SILog), mean absolute relative error (Abs Rel), mean squared relative error (Sq Rel), mean log10 error (log10), and root mean squared error (RMSE) to assess our approach. Waymo is not precisely a surround-view camera dataset because its camera layout contains around 120 degrees of missing angles. As a result, the primary ablation study experiments are performed on the nuScenes dataset. We also report the detection results on Waymo dataset and compare them to other approaches.

\begin{figure*}[h]
\includegraphics[width=1.0\textwidth]{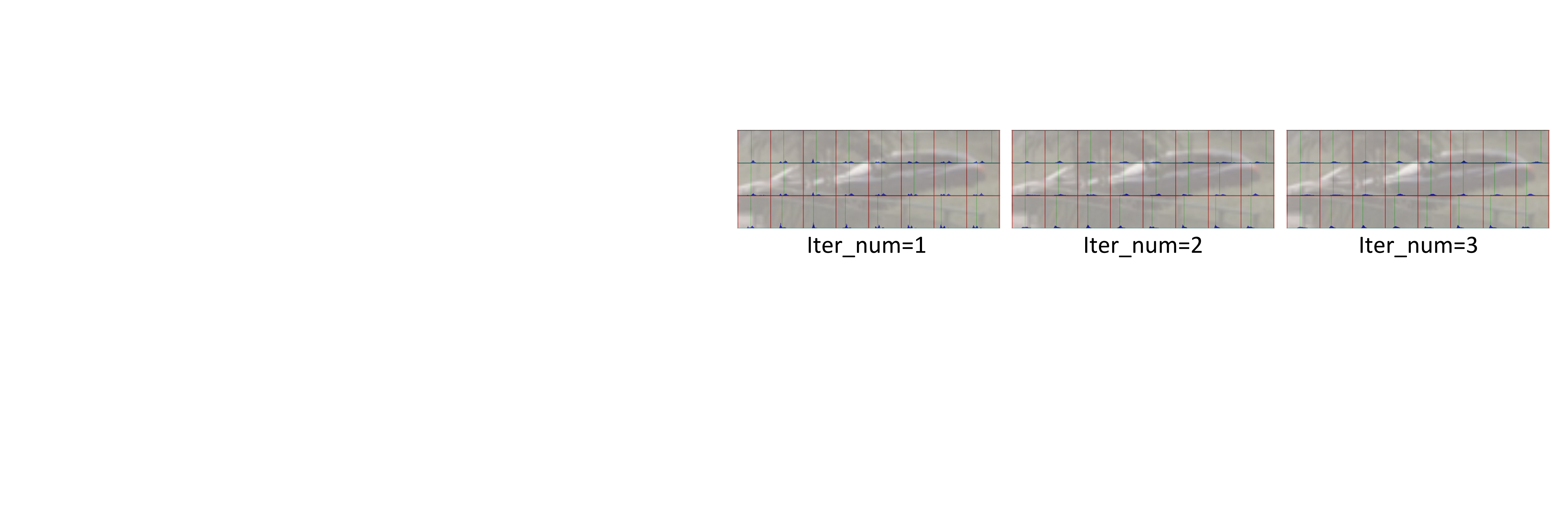}
\centering
\vspace{-0.3in}
\caption{Visualzation of depth prediction. The blue area is the distribution of depth prediction, while the green line represents the depth GT produced by the point cloud. Iter\_num denotes the number of iterations we apply during the inference time.}
\vspace{-0.1in}
\label{fig:visualize_iter}
\end{figure*}

\paragraph{Implementation details}
We implement BEVStereo++ based on BEVDepth~\cite{li2022bevdepth}. For motion compensation, we use a feature map with a downsampling rate of 4, which is subsequently used to construct temporal stereo. The depth feature's final form remains unchanged. To demonstrate the effectiveness of our method, we replace the depth module in BEVDepth with MVS~\cite{yao2018mvsnet} approach, which employs the same image resolution as our method. The learning rate is set to 2e-4, and the EMA technique and AdamW~\cite{adamw} are also used as the optimizer. During training, we use both image and BEV data augmentation.
\subsection{Analysis}
We perform numerous experiments to examine the mechanism of BEV-Stereo++ in order to better understand how it works. We choose BEVDepth ~\cite{li2022bevdepth}, SOLOFusion~\cite{park2022time} and BEVStereo(ours)~\cite{li2022bevstereo} as baseline, we also implement MVSNet~\cite{yao2018mvsnet} on BEVDepth as a comparison to  show the distinct benefit that BEVStereo++ provides,  detection results and recall results are used for comparison.

\paragraph{Memory analysis}
\begin{table}[t]
\centering
\caption{Memory usage and detection results of BEVDepth, BEVDepth with MVS and BEVStereo++. Frames denotes number of frames we use.}
\vspace{-0.1in}
\scalebox{0.95}{
\begin{tabular}{c|c|c|ccc}

\toprule
 \textbf{Method} & \textbf{Frames}&\textbf{Memory} &  \textbf{mAP}$\uparrow$  & \textbf{mATE}$\downarrow$ & \textbf{NDS}$\uparrow$ \\
\midrule
    BEVDepth & 2 & 6.49GB & 32.7 &  70.1 & 43.3  \\
  BEVDepth + MVS & 2 & 24.04GB & 34.7 & 67.1 & 44.9   \\
\midrule
  BEVStereo(ours) & 2 & 8.01GB & 34.6 & 65.3 & 45.3  \\
 BEVStereo++ & 2 & 8.44GB & 35.1 & 64.3 & 45.9   \\ 
 BEVStereo++ & 4 & 8.78GB & \textbf{37.8} & \textbf{62.7} & \textbf{48.3}   \\ 
\bottomrule
\end{tabular}}
\label{tab:memory-efficiency}
\end{table}

\begin{table}[t]
\centering
\vspace{-0.1in}
\caption{Evaluation of depth prediction on the nuScenes \emph{val} set.}
\vspace{-0.1in}
\scalebox{0.95}{
\begin{tabular}{l|ccccc} 
\toprule
\textbf{Method}& \textbf{SILog}$\downarrow$   & \textbf{Abs Rel}$\downarrow$ & \textbf{Sq Rel}$\downarrow$     & \textbf{log10}$\downarrow$  & \textbf{RMSE}$\downarrow$  \\
\midrule
   BEVDepth  & 21.740 & 0.155 & 1.223 & 0.060 & 5.269     \\
   BEVStereo(ours)  & 21.740 & 0.152 & 1.206 & 0.059 & 5.246     \\
 BEVStereo++ & \textbf{21.288} & \textbf{0.146} & \textbf{1.128} & \textbf{0.058} & \textbf{5.124}    \\
\bottomrule
\end{tabular}}
\label{tab:ablation-depth}
\end{table}

\begin{table}[t]
\centering
\vspace{-0.1in}
\caption{Recall results on the nuScenes \emph{val} set. Only boxes with velocity higher than 1m/s are maintained for analysis.  BEVDepth + MVS denotes replacing depth module in BEVDepth with MVS approach. Different thresholds are utilized depending on the distance between boxes' center.}
\vspace{-0.1in}
\scalebox{0.95}{
\begin{tabular}{l|cccc} 
\toprule
\textbf{Method} &\textbf{TH=0.5}   & \textbf{TH=1}& \textbf{TH=2}     & \textbf{TH=4}  \\
\midrule
   BEVDepth & 28.32 & 46.10 & 60.37 & 	71.18    \\
   BEVDepth + MVS & 27.67 & 46.40 & 59.99 & 71.26     \\
\midrule
 BEVStereo++ & \textbf{28.93} & \textbf{49.01} & \textbf{62.91} & \textbf{73.30}    \\
\bottomrule
\end{tabular}}
\label{tab:analysis-velocity}
\end{table}

\begin{table}[t]
\centering
\vspace{-0.1in}
\caption{Recall results on the nuScenes \emph{val} set. Only boxes with velocity lower than 1m/s are maintained for analysis.}
\vspace{-0.1in}
\scalebox{0.95}{
\begin{tabular}{l|ccccc} 
\toprule
\textbf{Method} &\textbf{TH=0.5}   & \textbf{TH=1}& \textbf{TH=2}     & \textbf{TH=4}  \\
\midrule
   BEVDepth & 32.80 & 53.58 & 70.00 & 80.89    \\
   BEVDepth + MVS & 33.61 & 54.23 & 69.89 & 80.57     \\
\midrule
 BEVStereo++ & \textbf{34.06} & \textbf{55.56} & \textbf{72.61} & \textbf{83.13}    \\
\bottomrule
\end{tabular}}

\label{tab:analysis-static}
\end{table}

\begin{table}[ht!]
\centering
\vspace{-0.1in}
\caption{Detection results on the nuScenes \emph{val} set. Only frames with ego vehicles moving at speeds less than 1 m/s are employed for evaluation.}
\vspace{-0.1in}
\scalebox{1.0}{
\begin{tabular}{c|ccc}
\toprule
 \textbf{Method}& \textbf{mAP}$\uparrow$  & \textbf{mATE}$\downarrow$ & \textbf{NDS}$\uparrow$ \\
\midrule
    BEVDepth &32.73 & 73.47 & 44.14  \\
  BEVDepth + MVS & 31.55 & 78.06 & 43.21   \\
\midrule
 BEVStereo++ & \textbf{33.94} & \textbf{65.66} & \textbf{46.59}
   \\ 
\bottomrule
\end{tabular}}
\label{tab:analysis-ego-vehicle}
\end{table}

\begin{table}[ht!]
\centering
\vspace{-0.1in}
\caption{Detection results on the nuScenes \emph{val} set. num\_iter denotes the number of iterations for $\mu$.}
\vspace{-0.1in}
\scalebox{0.95}{
\begin{tabular}{c|ccc}
\toprule
 \textbf{num\_iter} & \textbf{mAP}$\uparrow$  & \textbf{mATE}$\downarrow$ & \textbf{NDS}$\uparrow$ \\
\midrule
0 & 33.9 & 67.2 & 45.4  \\
1 & 34.3 & 66.6 & 45.7   \\ 
2 & 36.8 & 64.5 & 47.5  \\ 
3 & \textbf{37.8} & \textbf{62.7} & \textbf{48.3}  \\ 
\bottomrule
\end{tabular}}
\label{tab:iteration}
\end{table}

We keep track of memory usage and detection results to demonstrate how effectively we use our memory. We also monitor the same metrics for the MVS-based~\cite{yao2018mvsnet} approach for a fair comparison.

\begin{table*}[t!]
\centering
\caption{Detection results on the nuScenes \emph{val} set. Frames denotes number of frames we use. MC denotes Motion Compensation Module and WN denotes Weight Net.}
\vspace{-0.1in}
\scalebox{1.0}{
\begin{tabular}{l|c|c|c|ccccccc}
\toprule
 \textbf{Method} & \textbf{Frames} & \textbf{MC} & \textbf{WN} & \textbf{mAP}$\uparrow$  & \textbf{mATE}$\downarrow$ & \textbf{mASE}$\downarrow$  & \textbf{mAOE}$\downarrow$ & \textbf{mAVE}$\downarrow$ & \textbf{mAAE}$\downarrow$ & \textbf{NDS}$\uparrow$ \\
\midrule
    BEVDepth & 2 & - & - & 32.7 &  70.1 & 27.7 & 55.6 & 55.8 & 21.4 & 43.3  \\
\midrule
    SOLOFusion & 8 & - & - & 36.6 &  68.6 & - & - & 31.7 & - & 46.5  \\ 
    SOLOFusion & 16 & - & - & 37.7 &  65.5 & - & - & 30.7 & - & 47.4  \\
\midrule
    BEVStereo(ours) & 2& - & & 34.5 & 66.5 & 27.9 & 52.9 & 55.0 & 23.6 & 44.7   \\
 BEVStereo(ours) & 2 & - &\checkmark & 34.6 & 65.3 & 27.4 & 53.1 & 51.6 & 23.0 & 45.3    \\ \midrule
 BEVStereo++ & 2 & \checkmark & \checkmark & 35.1 & 64.3 & 28.1 & 53.3 & 47.7 & 22.7 & 45.9    \\
 BEVStereo++ & 4 & \checkmark & \checkmark & \textbf{37.8} & \textbf{62.7} & \textbf{28.1} & \textbf{52.6} & \textbf{40.1} & \textbf{22.4} & \textbf{48.3}    \\
\bottomrule
\end{tabular}}

\label{tab:performance-analysis-nuscenes}
\end{table*}

As illustrated in Tab.~\ref{tab:memory-efficiency}, BEVStereo++ increases the metrics on mAP, mATE, and NDS considerably at the expense of adding little memory consumption. When compared to using MVS~\cite{yao2018mvsnet} on BEVDepth~\cite{li2022bevdepth}, BEVStereo++ considerably reduces memory usage while boosting performance.

\paragraph{Performance analysis}

\begin{table}[t]
\centering
\vspace{-0.1in}
\caption{Detection results on the nuScenes \emph{val} set. Interval denotes the 
 number of interval frames between current key frame and the other frame we use.}
\vspace{-0.1in}
\scalebox{1.0}{
\begin{tabular}{c|ccc}
\toprule
 \textbf{Interval} & \textbf{mAP}$\uparrow$  & \textbf{mATE}$\downarrow$ & \textbf{NDS}$\uparrow$ \\
\midrule
  1 & 35.20 & 67.19 & 46.17  \\
  2 & 36.37 & 64.45 & 47.31   \\
  3 & \textbf{37.79} & 62.73 & \textbf{48.31} \\ 
  4 & 37.63 & 63.09 & 47.87 \\ 
  5 & 37.58 & \textbf{62.68} & 47.85 \\ 
\bottomrule
\end{tabular}}
\label{tab:analysis-long-sequence}
\end{table}

\begin{table}[t!]
\centering
\vspace{-0.1in}
\caption{Detection results on the nuScenes \emph{val} set. CA denotes class-agnostic. All results are conducted under the best hyper parameters.}
\vspace{-0.1in}
\scalebox{0.95}{
\begin{tabular}{l|c|ccc}
\toprule
 \textbf{Method} & CA & \textbf{mAP}$\uparrow$  & \textbf{mATE}$\downarrow$ & \textbf{NDS}$\uparrow$ \\
\midrule
    circle-nms & &  34.6 & 65.3 & 45.3  \\
    circle-nms & \checkmark&  24.9 & 80.6 & 38.0  \\
\midrule
  size-aware-circlenms & & \textbf{35.1} & 64.7 & \textbf{45.6}   \\ 
  size-aware-circlenms & \checkmark&   33.3 & \textbf{64.1} & 45.0 \\ 
\bottomrule
\end{tabular}}
\label{tab:circlenms}
\end{table}

\begin{table}[t!]
\centering
\vspace{-0.1in}
\caption{Detection results on the Waymo \emph{val} set. Only 1/5 data are used for training. $\dagger$ denotes pretrained with FCOS3D++.}
\vspace{-0.1in}
\scalebox{0.95}{
\begin{tabular}{l|ccc}
\toprule
 \textbf{Method}& \textbf{mAP}$\uparrow$  & \textbf{mAPH}$\uparrow$ & \textbf{mAPL}$\uparrow$  \\
\midrule
BEVDepth-R50~\cite{li2022bevdepth} & 38.47 & 33.36 & 27.34  \\ 
MV-FCOS3D++-R101~\cite{wang2022mv}$\dagger$ & 46.65 & \textbf{44.25} & 33.80  \\ 
\midrule
BEVStereo++-R50 &  40.08 & 34.60 & 28.25   \\
BEVStereo++-R101 & \textbf{49.64} & 43.65 & \textbf{35.76} \\
\bottomrule
\end{tabular}}
\label{tab:performance-analysis-waymo}
\end{table}



To begin with, we demonstrate the performance comparison under the nuScenes~\cite{caesar2020nuscenes} evaluation metrics. As shown in Tab.~\ref{tab:performance-analysis-nuscenes}, Our BEVStereo++ outperforms BEVDepth on mAP, mATE and NDS. Tab.~\ref{tab:ablation-depth} shows that the accuracy of depth estimation is improved by introducing our design. As shown in Tab.~\ref{tab:performance-analysis-waymo}, our experiments on Waymo Open dataset also draw the same conclusion.

We assess the performance of BEVStereo++ under challenging conditions such as moving objects, and static ego vehicles in order to show how well it adapts to complicated outdoor environments. Tab.~\ref{tab:analysis-velocity} demonstrates that BEVStereo++ still has the ability to improve performance even while MVS approach fails when dealing with moving objects. The static objects, which make up the majority of MVS schema's contribution, are also used to evaluate our method. As shown in Tab.~\ref{tab:analysis-static}, BEVStereo++'s ability of perceiving static objects is even higher than BEVDepth with MVS. We choose frames whose ego vehicle has a low velocity for evaluation since MVS cannot handle situations when this occurs. As can be seen in Tab.~\ref{tab:analysis-ego-vehicle}, BEVStereo++ still improves performance even when MVS fails in these conditions. It is important to note that BEVStereo++ still produces the similar results when faced with circumstances like moving objects and static ego vehicles if $\mu$ is not updated during the inference step. This demonstrates that our schema is capable of guiding the Depth Module to produce better $\mu$ and maintaining the initial prediction of $\mu$ in the face of these eventualities.

\subsection{Ablation Study}

\paragraph{Iteration of $\mu$ and $\sigma$}

We conduct various experiments during the inference stage by modifying the number of iterations just to verify the function of iterating $\mu$ and $\sigma$. As illustrated in Tab.~\ref{tab:iteration}, the detection results improve as the number of iterations grows. The visualization result in Fig.~\ref{fig:visualize_iter} draws the same conclusion.

\paragraph{Weight Net}
We run the experiments under identical conditions without Weight Net to assess its validity. Weight Net promotes the detection results, as shown in Tab.~\ref{tab:performance-analysis-nuscenes}.

\paragraph{Frame Fusion}

To explore the function of our frame fusion schema, we conduct several experiments under different settings. All the experiments use four frames, but the frame index that we use are different. As illustrated in Tab.~\ref{tab:analysis-long-sequence}, we can observe that when we use further frames, we can acquire better performance. However, the performance reaches its peak when we set the frame index to 3.

\begin{figure}[t]
\vspace{-0.1in}
\includegraphics[width=0.45\textwidth]{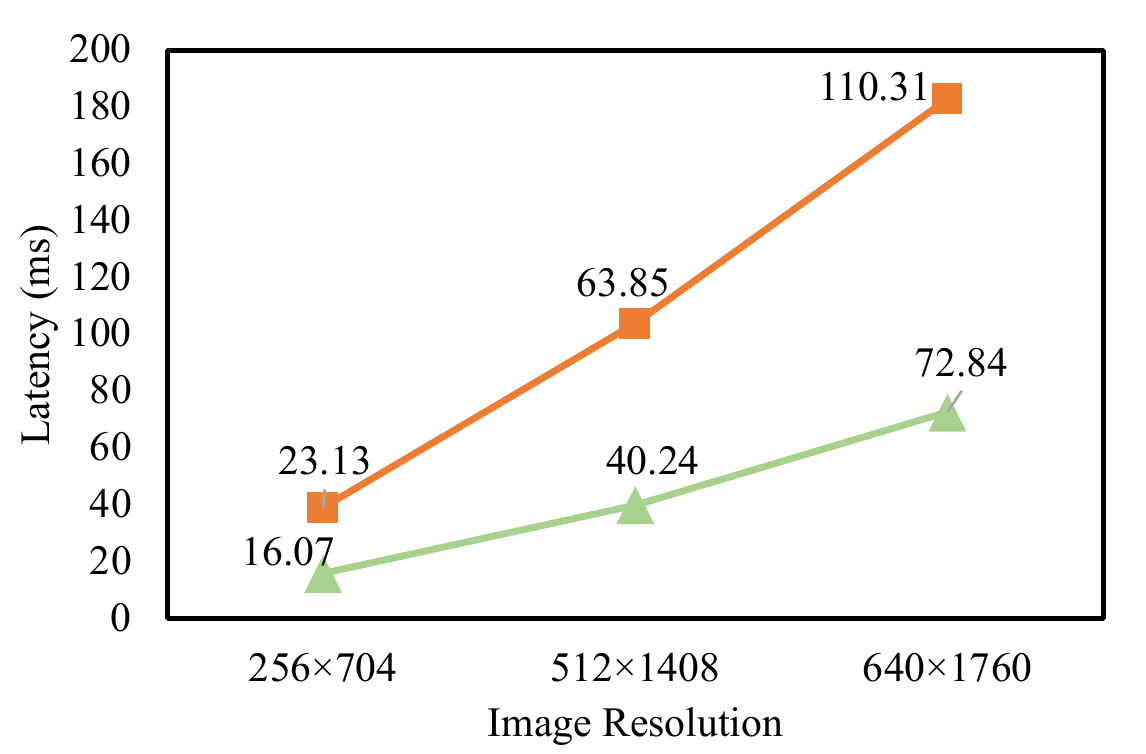}
\centering
\vspace{-0.1in}
\caption{Latency comparison between different frame fusion schemas. The orange line reflects the reimplemented frame fusion schema from SOLOFusion~\cite{park2022time} and the green line represents our frame fusion schema. During the inference time, we count the latency of 50 frames, timing begins after warm-up.}
\label{fig:frame_fusion_latency}
\vspace{-0.2in}
\end{figure}

\begin{figure}[t]
\includegraphics[width=0.45\textwidth]{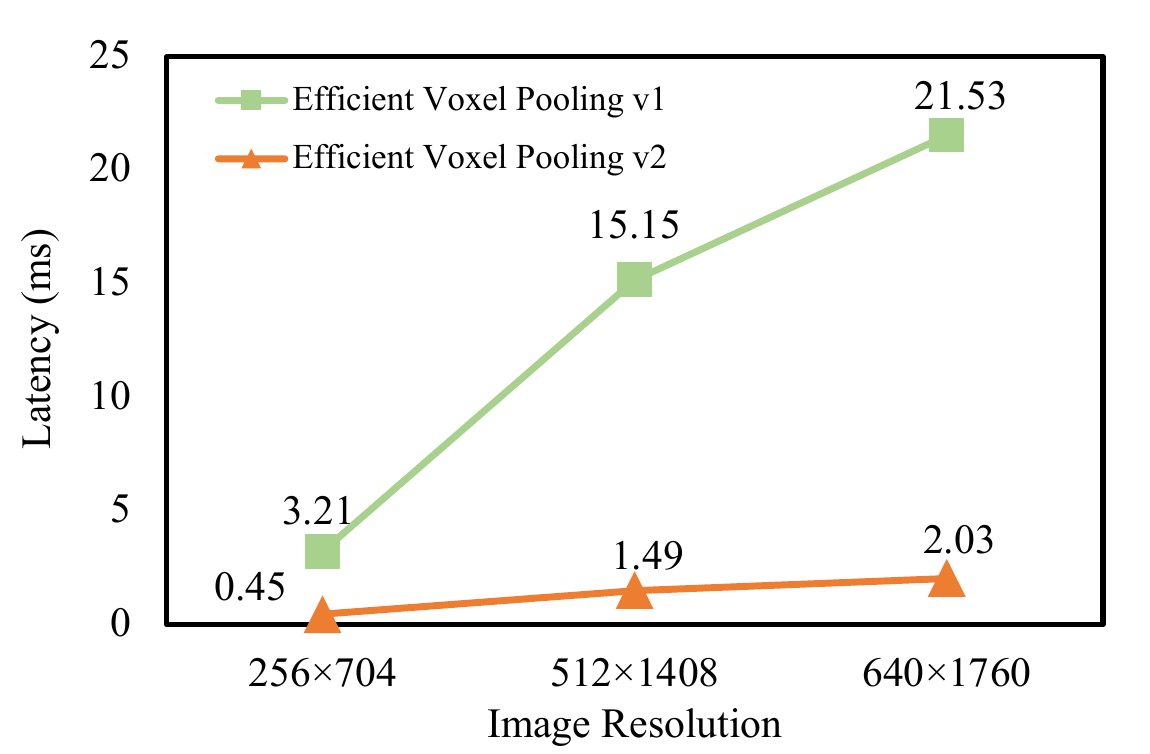}
\centering
\vspace{-0.1in}
\caption{Latency comparison between different frame fusion schemas. The orange line reflects the reimplemented frame fusion schema from SOLOFusion~\cite{park2022time} and blue line represents our frame fusion schema. During the inference time, we count the latency of 50 frames, timing begins after warm-up.}
\label{fig:efficient-voxel-pooling}
\vspace{-0.2in}
\end{figure}

To test the high efficiency of our frame fusion schema, we compare our frame fusion schema with SOLOFusion~\cite{park2022time}. As shown in Fig.~\ref{fig:frame_fusion_latency}, our schema offers faster inference speed while maintaining higher performance.
\paragraph{Size-aware Circle NMS}

\begin{figure*}[t!]
\includegraphics[width=1.0\textwidth]{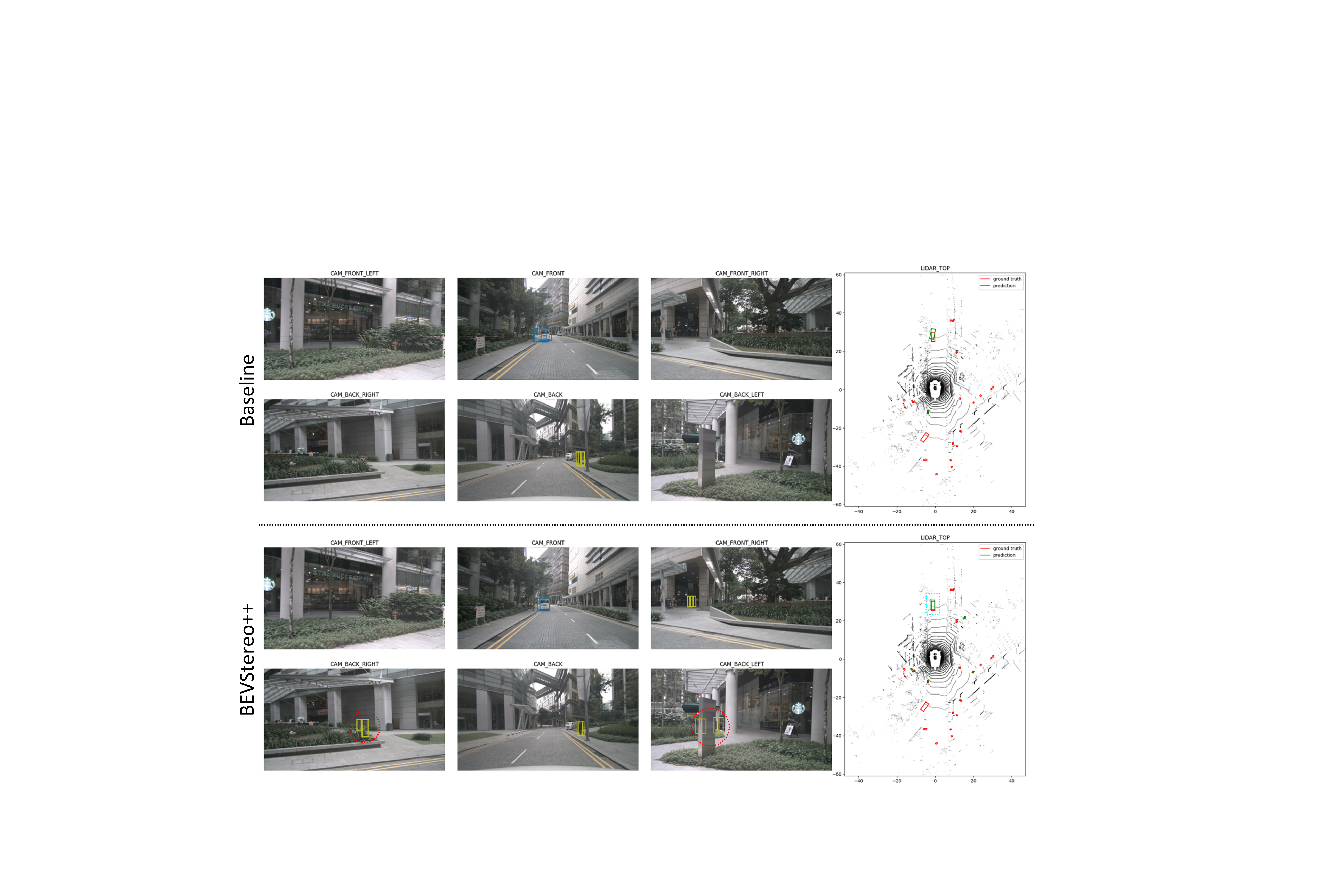}
\centering
\vspace{-0.2in}
\caption{Visualization of detection results. The blue dotted rectangle designates the object recognized by our approach is more accurate on localization, while the red dotted circle designates the object detected by BEVStereo++ but missed by the baseline.}
\label{fig:visualize_detection}
\end{figure*}

\subsection{Visualization}
\begin{figure*}[t]
\includegraphics[width=0.98\textwidth]{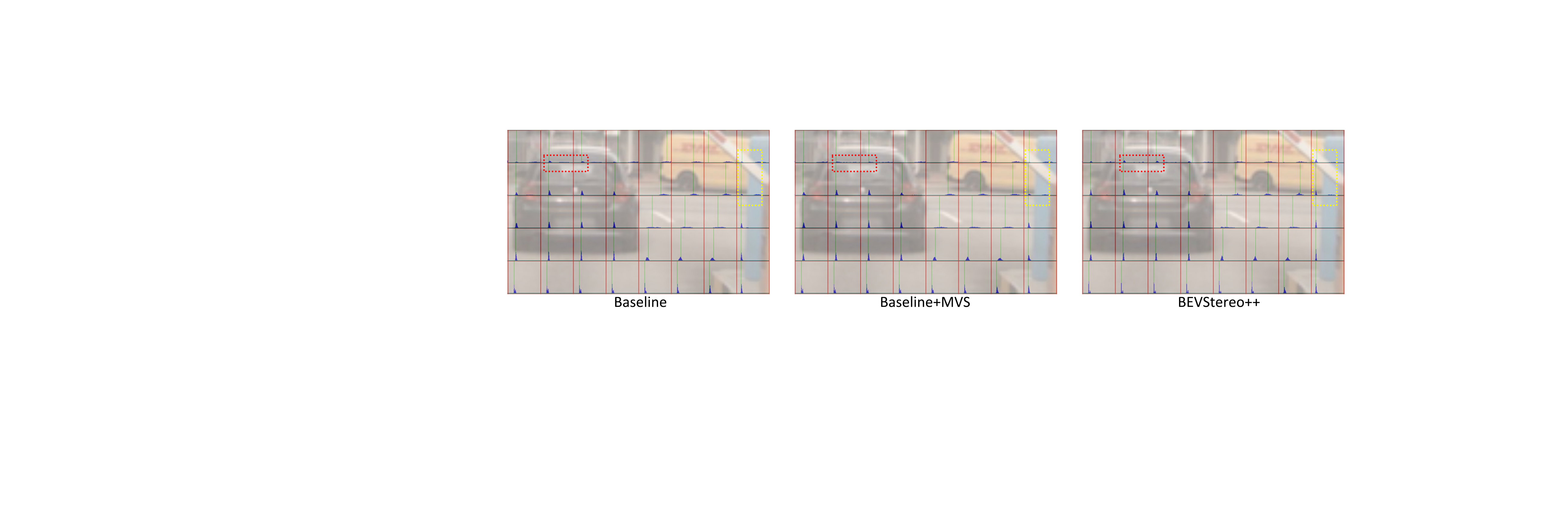}
\centering
\caption{Visualization of depth prediction. we take previous SOTA model BEVdepth as the baseline. The blue area is the distribution of depth prediction, while the green line represents the depth GT produced by the point cloud. The red dotted boxes denotes the promotion of depth prediction on moving objects and the yellow dotted boxes denotes the promotion of depth prediction on static objects.}
\label{fig:visualize_depth}
\end{figure*}

As illustrated in Fig.~\ref{fig:visualize_depth}, we can find that BEVStereo++ has the ability to promote the accuracy of depth estimation on both moving and static objects. Fig.~\ref{fig:visualize_detection} also shows the visualization of the detection results, and demonstrates the performance promotion brought by BEVStereo++.

\begin{table*}[h]
\centering
\caption{Comparison on the nuScenes \emph{test} set. L denotes LiDAR and C denotes camera.}
\resizebox{0.9\textwidth}{!}{
\begin{tabular}{l|c|cccccc|c}
\toprule
\textbf{Method}                                   & \textbf{Modality} & \textbf{mAP}$\uparrow$  & \textbf{mATE}$\downarrow$ & \textbf{mASE}$\downarrow$  & \textbf{mAOE}$\downarrow$ & \textbf{mAVE}$\downarrow$ & \textbf{mAAE}$\downarrow$ & \textbf{NDS}$\uparrow$ \\
\midrule
CenterPoint                         & L        & 0.564 & - & - & -   & - & - & 0.648  \\ 
\midrule
FCOS3D~\cite{wang2021fcos3d}                                   & C        & 0.358 & 0.690 & 0.249 & 0.452   & 1.434 & 0.124 & 0.428  \\
DETR3D~\cite{wang2022detr3d}                                   & C        & 0.412 & 0.641 & 0.255 & 0.394   & 0.845 & 0.133 & 0.479  \\
BEVDet-Pure~\cite{huang2021bevdet}                              & C        & 0.398 & 0.556 & 0.239 & 0.414   & 1.010 & 0.153 & 0.463  \\
BEVDet-Beta                              & C        & 0.422 & 0.529 & \textbf{0.236} & 0.396   & 0.979 & 0.152 & 0.482  \\
PETR~\cite{liu2022petr} & C        & 0.434 & 0.641 & 0.248 & 0.437   & 0.894 & 0.143 & 0.481  \\
PETR-e                                   & C        & 0.441 & 0.593 & 0.249 & 0.384   & 0.808 & 0.132 & 0.504  \\
BEVDet4D~\cite{huang2022bevdet4d}                                 & C        & 0.451 & 0.511 & 0.241 & 0.386   & 0.301 & 0.121 & 0.569  \\
BEVFormer~\cite{li2022bevformer}                                & C        & 0.481 & 0.582 & 0.256 & 0.375   & 0.378 & 0.126 & 0.569  \\ 
PETRv2~\cite{liu2022petrv2}                                & C        & 0.490 & 0.561 & 0.243 & \textbf{0.361}   & 0.343 & \textbf{0.120} & 0.582  \\ 

BEVDepth~\cite{li2022bevdepth}                               & C        & 0.503 & 0.445 & 0.245 & 0.378 & 0.320 & 0.126 & 0.600 \\
SOLOFusion~\cite{park2022time} & C & 0.540 & 0.453 & 0.257 & 0.376 & 0.276 & 0.148 & 0.619 \\
\midrule
BEVStereo(ours)                                 & C        & 0.525 & 0.431 & 0.246 & 0.358 & 0.357 & 0.138 & 0.610 \\
BEVStereo++ & C & \textbf{0.546} & \textbf{0.427} & 0.253 & 0.412 & \textbf{0.251} & 0.132 & \textbf{0.625} \\

\bottomrule
\end{tabular}}
\label{tab:test}
\end{table*}

We compare BEVStereo++ with the size-aware circle NMS to BEVStereo++ with the conventional circle NMS as our baseline. They are subjected to class-aware and class-agnostic procedures in order to test the validity of size-aware circle NMS.

As shown in Tab.~\ref{tab:circlenms}, our size-aware circle NMS improves on the matrices of mAP, mATE, and NDS when using class-aware NMS. The traditional distance-based circle NMS has completely lost its capacity to suppress under class-agnostic circumstance, while our size-aware circle NMS continues to function well.

We utilize latency as a measure to show how Efficient Voxel Pooling V2 improves performance. We measure the module's latency for both Efficient Voxel Pooling V1(used in BEVDepth) and Efficient Voxel Pooling V2 during the inference time. For a fair comparison, both experiments are carried on the same device with the same batch size. We can see from Fig.~\ref{fig:efficient-voxel-pooling} that our Efficient Voxel Pooling V2 has demonstrated significant superiority at various resolutions.

\begin{table}[t]
\centering
\caption{Comparison on the nuScenes \emph{val} set. L denotes LiDAR and C denotes the camera.}
\scalebox{1.10}{
\setlength{\tabcolsep}{1.5pt}
\begin{tabular}{l|c|c|cc} 
\toprule
\textbf{Method}             & \textbf{Resolution} & \textbf{Modality} & \textbf{mAP}$\uparrow$  & \textbf{NDS}$\uparrow$ \\
\midrule
CenterPoint-Voxel~\cite{yin2021center}  &         -      & L        & 56.4   & 64.8  \\
CenterPoint-Pillar &       -        & L        & 50.3 & 60.2  \\ 
\midrule
FCOS3D~\cite{wang2021fcos3d}            & 900$\times$1600      & C        & 29.5 & 37.2  \\
DETR3D~\cite{wang2022detr3d}             & 900$\times$1600      & C        & 30.3 & 37.4  \\
BEVDet-R50~\cite{huang2021bevdet}         & 256$\times$704       & C        & 28.6 & 37.2  \\
BEVDet-Base        & 512$\times$1408      & C        & 34.9 & 41.7  \\
PETR-R50~\cite{liu2022petr}           & 384$\times$1056      & C        & 31.3 & 38.1  \\
PETR-R101          & 512$\times$1408      & C        & 35.7 & 42.1  \\
PETR-Tiny          & 512$\times$1408      & C        & 36.1 & 43.1  \\
BEVDet4D-Tiny~\cite{huang2022bevdet4d}      & 256$\times$704       & C        & 32.3 & 45.3  \\
BEVDet4D-Base      & 640$\times$1600      & C        & 39.6 & 51.5  \\
BEVFormer-S~\cite{li2022bevformer}        &       -        & C        & 37.5 & 44.8  \\
BEVDepth-R50~\cite{li2022bevdepth}        & 256$\times$704       & C        & 35.9 & 48.0  \\
BEVDepth-ConvNext       & 512$\times$1408      & C        & 46.2 & 55.8  \\
\midrule
BEVStereo-R50(ours)~\cite{li2022bevdepth}& 256$\times$704  & C        & 37.6 & 49.7\\
BEVStereo-ConvNext(ours)   & 512$\times$1408       & C        & 47.8 & 57.5  \\
\midrule
BEVStereo++-R50        & 256$\times$704       & C        & 41.1 & 53.2  \\
BEVStereo++-ConvNext   & 512$\times$1408       & C        & \textbf{49.1} & \textbf{58.7}  \\
\bottomrule
\end{tabular}}
\label{tab:val}
\end{table}

\subsection{Benchmark Result}

We compare BEVStereo++ with other SOTA methods like CenterPoint ~\cite{yin2021center}, FCOS3D~\cite{wang2021fcos3d}, DETR3D~\cite{wang2022detr3d}, BEVDet~\cite{huang2021bevdet}, PETR~\cite{liu2022petr}, BEV-Det4D~\cite{huang2022bevdet4d}, BEVFormer~\cite{li2022bevformer}, BEVStereo(ours)~\cite{li2022bevstereo} and SOLOFusion~\cite{park2022time}. We evaluate our BEVStereo++ on the nuScenes \emph{val} and \emph{test} set. As shown in Tab.~\ref{tab:val} and Tab.~\ref{tab:test}, BEVStereo++ achieves the highest score of camera-based methods on both mAP and NDS. It is worth mentioning that we choose the SOTA LiDAR-based method, the CenterPoint method~\cite{yin2021center}, as the upper bound. The BEVStereo++ significantly improves the performance of the camera-based method and its performance is close to the LiDAR-based method.

\section{Conclusion}
This paper proposes a novel multi-view object detector, namely BEVStereo++. BEVStereo++ improves performance without significantly increasing memory usage by applying a dynamic temporal stereo technique. BEVStereo++ also avoids failure at reasoning the depth of moving objects and static ego vehicles by designing a Motion Compensation Module. In addition, we propose size-aware circle NMS, which considers the size of boxes while avoiding the laborious computation of rotated IoU. Under both class-aware and class-agnostic circumstances, our size-aware circle NMS shows superior performance. Last but not least, we present Efficient Voxel Pooling v2, which speeds up voxel pooling by improving memory efficiency.

\bibliographystyle{IEEEtran}
\bibliography{IEEEabrv,ref}

\end{document}